\documentclass[journal]{IEEEtran}
\usepackage{hyperref,graphicx}
\usepackage{ifpdf}

\usepackage{cite}

\ifCLASSINFOpdf
\else
\fi
\usepackage{amsmath}
\usepackage[symbol]{footmisc}

\usepackage{algorithmic}

\usepackage{adjustbox}
\usepackage{caption}    
\usepackage{tabularx}
\usepackage[dvipsnames]{xcolor}

\usepackage{url}

\begin{document}
%
\title{Class-Level Feature Selection Method Using Feature Weighted Growing Self-Organising Maps}

\author{\IEEEauthorblockN{Andrew Starkey, Uduak Idio Akpan$^{\dagger}$, Omaimah AL Hosni and Yaseen Pullissery}
\IEEEauthorblockA{\\ {School of Engineering}
University of Aberdeen, Aberdeen, UK\\
$^{\dagger}$Akwa Ibom State University, Nigeria\\
(a.starkey, u.akpan.18)@abdn.ac.uk: uduakidio@aksu.edu.ng\\
}}


%


\maketitle

\begin{abstract}
There have been several attempts to develop Feature Selection (FS) algorithms capable of identifying features that are relevant in a dataset. Although in certain applications the FS algorithms can be seen to be successful, they have similar basic limitations. In all cases, the global feature selection algorithms seek to select features that are relevant and common to all classes of the dataset. This is a major limitation since there could be features that are specifically useful for a particular class while irrelevant for other classes, and full explanation of the relationship at class level therefore cannot be determined. While the inclusion of such features for all classes could cause improved predictive ability for the relevant class, the same features could be problematic for other classes. In this paper, we examine this issue and also develop a class-level feature selection method called the Feature Weighted Growing Self-Organising Map (FWGSOM). The proposed method carries out feature analysis at class level which enhances its ability to identify relevant features for each class. Results from experiments indicate that our method performs better than other methods, gives explainable results at class level, and has a low computational footprint when compared to other methods. 
\end{abstract}



%
\IEEEpeerreviewmaketitle

\section{Introduction}
As the world is giving more attention to data exploration and extraction, and also the computational cost of this process and subsequent modelling, there is an urgent need to have algorithms that can extract and/or select the features that are relevant and useful in the datasets automatically and to be capable of explaining these relationships. Feature selection (FS) is the process of finding the features that are useful in a dataset. FS methods are most often employed at the data pre-processing stage. This helps to achieve efficient data reduction, find accurate data models, and improve predictive accuracy of classifiers, as well as to reduce their computational complexity. As the size of datasets keeps growing by the day, both in respect to the number of features and samples, there have been several attempts to develop FS algorithms to achieve these tasks. Although in certain applications the FS algorithms can be seen to be successful, they have similar basic limitations. For example, some of the FS algorithms are capable of correctly selecting the important features only in cases where the importance of the features is uniform or common for all the classes.
However, and regrettably, these algorithms often fail in correctly identifying and selecting the important features in datasets where the input feature relevance is different or not uniform i.e. where a feature could be relevant for one class and irrelevant or noisy for another class. The failure of the FS algorithms to identify this type of feature relevance is due to the inability of the FS algorithms to carry out feature relevance analysis at class level. In all cases, the global FS algorithms seek to select features relevant and common to all classes of the dataset \cite{janzing2020feature}\cite{pourpanah2019feature}. This is a major limitation because there could be features that are specifically useful for a particular class while irrelevant for other classes. While the inclusion of such features for all classes could result in improved predictive ability for the relevant class, some of those features could be problematic for other classes. Due to the recent improvements in AI with Deep Learning the inability of the FS algorithms to perform this task is masked by the superior classification capabilities of the Deep Learning algorithms, which are able to automatically determine the relevant features for a class within their hidden layers but in a manner that is not explainable to the end user.
Another problem associated with these global FS techniques, and similar to the black box nature of the classifiers, is that these features cannot be linked to any specific prediction output, and therefore they cannot be seen to provide transparency or explainability to the process. In \cite{dovsilovic2018explainable} and \cite{liu2019incorporating}
this Explainable Artificial Intelligence (XAI) technique (which implies the process of associating a set of attributes to a prediction outcome) is described as feature attribution. In order to overcome the limitation of global FS methods, there is a need for the FS to be carried out for each class. This is considered as local FS method because the relevant features are identified for each class in the dataset. With this, it is possible to interpret or explain the rationale for each predicted class by linking the decision back to those features identified for the class \cite{dovsilovic2018explainable} and \cite{liu2019incorporating}. In this paper, we develop a class-level FS method called the Feature Weighted Growing Self-Organising Map (FWGSOM). This method utilises information from a Growing Self-Organising Map (GSOM) \cite{alahakoon2000dynamic}. The information from a trained GSOM is used to guide the process of feature analysis at both the class and node level. While we observe that the most common metric for FS techniques evaluation is based on improvement in classification accuracy when the selected features are used, we however argue that this may not always be the best metric to determine if the FS technique was successful in identifying the relevant features. For instance, deep neural networks or deep learning (DL) has been seen in many cases to provide superior classification accuracies for many datasets with irrelevant features, as the method in theory automatically undertakes its own internal feature selection process. However, the DL process is considered as “black box” because it lacks transparency and explainability \cite{hofmann2003rule} \cite{biryulev2010research}. Furthermore, DL merely suppresses the noisy features during classification and does not provide information about which features are relevant for which class. Despite the ability of DL to provide the best classification results, we argue that for some domains, the knowledge of what is important in the datasets is as important as the classification or predictive accuracy. With increased scrutiny in AI models and the results obtained from their opaque nature, and with the introduction of legislation restricting the use of AI in many countries around the world, the justification for methods that give understanding of the data that drives AI models has never been more important \cite{lyon2020us} \cite{elisa2020us} \cite{misha2021us}.  Previous research by the authors \cite{akpan2021review} conducted a study of the performance of different classification methods on different data related problems. As would be expected, the DL was observed to have performed better than other classification techniques in most of the data related problems under investigation, but is unexplainable and has high computational cost. In contrast, the research presented in this paper focuses on presenting a transparent and explainable FS algorithm, low in computational cost, that can identify and present the features that are relevant for each class. This in turn will ultimately cause improvement in the classification process, independent of which algorithm is used. Therefore, this paper presents a class-based feature selection method inspired by GSOM and compares its performance and computational complexity against other common FS algorithms for a range of data problems. More detail of the algorithm will be presented in the following sections.
\section{Background Information}
\subsection{Feature selection}
There are generally three categories of FS methods viz; Filter, wrapper and embedded methods. 
\subsubsection{Filter methods}
The filter feature selection methods are independent of the learning algorithm. They rely on a function for evaluating which features are important or irrelevant to the classification process. Filter methods adopt statistical measures to rank available input features. The feature with lower predetermined score threshold is rejected automatically by the algorithm \cite{dovsilovic2018explainable}. These methods select a subset of features by means of ranking of the values of all features on the evaluation function. The determination of which features are relevant is based on the correlation between predictor variables and the targets. There are several criteria for feature ranking in the filter FS method. The Correlation Based Feature Selection (CFS) calculates the coefficients of the input features by using the target index and measuring the linear correlation between the features. In this process, the input features with the highest correlation coefficients are regarded as the most relevant and are selected accordingly. The most commonly used CFS is the Pearson’s correlation coefficient score (CORR) and the correlation coefficient $\rho$ is computed by the expression in \eqref{eq:1}. 

\begin{equation}
	\rho=\frac{\sum_{i}(x_{i}-\bar{x})(y_{i}-\bar{y})}{\sqrt{\sum_{i}(x_{i}-\bar{x})^{2}}\sqrt{\sum_{i}(y_{i}-\bar{y})^{2}} }
	\label{eq:1}
\end{equation}

Where $x$ and $y$ are the variables and $\bar{x}$ and $\bar{y}$ are the mean of $x$ and $y$ respectively. Practically, $-$1 $\leq$ $\rho$ $\leq$ 1 with $\rho$ $=1$ or $-$1, representing complete positive and negative correlation respectively while $\rho$ $=0$ depicts no correlation between the variables. A major advantage of CFS is the elimination of input features with correlation coefficients ($\rho$) of Zero or very close to zero known as near zero correlation to classes. In real practice, as demonstrated by \cite{hofmann2003rule}, CFS often fails in some real-world datasets due to the non-linear correlation of the variables. 
Nearest neighbour ranking criterion such as Relief and ReliefF are also very common. Rather than search through feature combinations, the Relief and ReliefF apply the concept of nearest neighbours to extract feature statistics which indirectly are responsible for the interactions. Relief calculates proxy statistics (feature weights) for the individual features and uses these to estimate the quality of the feature and their relevance to the target concept (i.e. predicting endpoint value). Relief has its strength on being able to estimate the quality of a feature in the context of other features and makes no assumptions concerning the sample size of the dataset. Unfortunately, the Relief technique’s performance deteriorates as the irrelevant features become large \cite{hofmann2003rule}.
\subsubsection{Mutual Information}
Mutual information (MI) is a univariate filter method. It calculates the relevancy of each feature based on Information theory \cite{biryulev2010research} \cite{kotthoff2016algorithm}. According to Shannon, entropy can be defined in \eqref{eq:2}.
\begin{equation}
	H(Y)=-\sum \rho(y)log(\rho(y))
	\label{eq:2}
\end{equation}

Equation 2 measures the uncertainty (information content) of the random variable $Y$. The uncertainty is related to the probability of occurrence of an event. The conditional entropy of $Y$ given a variable $X$ is expressed in \eqref{eq:3}.
\begin{equation}
	H(Y|X)=-\sum_{x} \sum_{y} \rho(X,y)log(\rho(y|X))
	\label{eq:3}
\end{equation}

The joint entropy of Y and X is the sum of uncertainty contained by the two variables. Equation 3 implies that observing a variable X, the uncertainty of Y is reduced. The decrease in uncertainty is given in \eqref{eq:4}.

\begin{equation}
	I(Y,X)= H(Y)-H(Y|X)
	\label{eq:4}
\end{equation}

Mutual Information calculates the amount of information that one random variable has about another variable. This helps in identifying the features that are relevant to the label or target of the data, MI formal definition can be expressed in equation \eqref{eq:5}. 
\begin{equation}
	MI(x;y)=\sum_{i=0}^{n}\sum_{jj=0}^{n} \rho(x(i),y(j))log(\frac{\rho(x(i),y(j))}{\rho(x(i).\rho(y(j)})
	\label{eq:5}
\end{equation}

Where MI is zero when $x$ and $y$ are statistically independent. 

\subsubsection{F-Score method}
F-Score is a simple feature selection filter method achieved by evaluating the discrimination of two sets of real values. The F-score uses F-distribution to determine feature relevance and contributions to classification accuracy of the classifier. The F-score method looks at the relationship between each descriptive feature and the target feature using the F-distribution. Given training vectors $x_{k}$, $k$ = 1, 2 ,..., m, if the number of positive and negative instances are $n_{+}$ and $n_{-}$ respectively, then the F-score of the $i^{th}$ factor is defined in \eqref{eq:6}, as presented in \cite{ding2009feature}.

\begin{figure*}
\begin{equation}
	F(i)=\frac{({\bar{x_{i}}^{(+)}}-\bar{x_{i}})^{2}+({\bar{x_{i}}^{(-)}}-\bar{x_{i}})^{2}}{{\frac{1}{n_{+}-1}}\sum_{k=1}^{n+}({\bar{x}_{k,i}}^{(+)}-{\bar{x}_{i}}^{(+)})^{2}+{\frac{1}{n_{-}-1}}\sum_{k=1}^{n-}({\bar{x}_{k,i}}^{(-)}-{\bar{x}_{i}}^{(-)})^{2}}
	\label{eq:6}
\end{equation}
\end{figure*}

Once the F-score values are calculated for all features, a threshold value is calculated by taking the mean of all F-score values. The threshold value is used as a criterion to select the relevant features. This is done by selecting the features whose F-score values are greater than the threshold. The features with F-score values less than the threshold value are discarded. The larger the F-score, the more discriminative the feature is assumed to be. One of the disadvantages of F-score method is that it does not take the mutual information between features into account \cite{ding2009feature}.

\subsection{Wrapper Methods}
Contrary to the filter methods which employ no learning algorithm for evaluation, the wrapper methods involve specific learning algorithms usually adopted to evaluate the performance and accuracy of the selected features. Feature subsets which gives better solutions are considered relevant and accepted \cite{chandrashekar2014survey} \cite{zongker1996algorithms}. However, the deployment of the wrapper methods is difficult especially with high dimensional datasets which pose a big computational burden even for the simplest learning algorithm. Common search algorithms used for wrapper feature selection include the branch and bound method \cite{kohavi1997wrappers} \cite{houstis2000future} and other metaheuristic algorithms. Also common, are sequential selection based algorithms such as the sequential forward selection (SFS), sequential backward selection (SBS), sequential forward floating Selection (SFFS) and sequential backward floating selection (SBFS). 
The SFS and SFFS algorithms usually begin with an empty set and add input features individually and sequentially as they calculate the performance with cross validation at each round. This procedure is continued iteratively until convergence. The SFFS, an extension of the SFS overcomes the lack of backtracking associated with SFS. The SFFS backtracks (performs backward steps), after each forward step, if the objective function increases, that is if the resulting subsets have better performance than the ones previously evaluated at that level \cite{guyon2003introduction}. The SBS operates in reverse direction to the SFS. Unlike the SFS that begins with an empty set, the SBS starts with a set of all the inputs features. The SBS sequentially eliminates the feature whose removal from the set, results in the best score and considers it as irrelevant \cite{zhao2010advancing} \cite{aha1996comparative} \cite{li2004comparative}. The input feature with the least performance accuracy is considered irrelevant and is removed. This process is performed iteratively until convergence, where improvement of classification performance is no longer possible by removing any of the remaining input features. As presented in \cite{bertolazzi2016integer} unlike the SFS which has no ability to backtrack, the SBS can backtrack which allows for re-addition of earlier eliminated features. 

\subsubsection{Heuristic Search Algorithms}
Heuristic Search algorithms are less computationally demanding than the exhaustive search algorithms. The algorithms work in two phases. In the first phase, the features are ranked according to a selected evaluation measure. In the second phase, an iterative process passes through the ranked features from the first to the last. The classification accuracy is obtained with the first ranked feature in the list. In the second step, the classifier is run with the first two features of the list, and a paired t-test is performed to determine the statistical significance degree of the differences. If it is lower than a threshold, the second feature is ignored and the processes repeats the second step on the third ranked feature in the list. In this process, only the features that increase classification accuracy are added. At the end of the second phase, a list of selected features will be provided \cite{bertolazzi2016integer} \cite{mehri2018comparative}.

\subsubsection{Genetic algorithm}
Genetic algorithm (GA) is a very adaptive and efficient method for feature selection. It provides the flexibility for data mining experts to change the configuration of the algorithm to improve the results. GA is an optimization technique. GA is one of the heuristic search algorithms that mimic the natural evolution process \cite{mostert2019insights} \cite{smith2009cross}. 

\subsubsection{Recursive feature elimination and cross validation}
The Recursive Feature Elimination and Cross Validation (RFECV) explores features recursively considering smaller subsets of features. The algorithm ranks the feature subsets as opposed to feature ranking. In an iterative procedure, the classifier is trained, and weights are assigned, ranking criterion for all features are calculated and the features with smallest ranking criterion are removed first. The procedure is repeated until an optimal number of features is achieved. This optimal number of features are chosen with the help of cross-validation. The SVM classifier is the most commonly used classifier with RFECV feature selection method \cite{lindauer2015autofolio} \cite{sakamoto1986akaike}. 

\subsection{Embedded method}
The embedded method was conceived in attempt to integrate the strengths of the filter and wrapper methods for better solutions \cite{chandrashekar2014survey} \cite{zongker1996algorithms}. Unlike the ﬁlter and wrapper approaches, the embedded methods are incorporated into the learning algorithm for selection of relevant features during training and are specific to the learning algorithm. In the embedded methods, the training and feature selection parts are inseparable. In summary, the feature selection technique is built into the classifier in the embedded methods.

\subsubsection{Least Absolute Shrinkage and Selection Operator}
The Least Absolute Shrinkage and Selection Operator (LASSO) technique is a powerful method that performs regularization and feature selection. The method applies regularization (shrinking) in order to keep the sum of the absolute values of the model parameters to be less than a fixed value (upper bound). The regularization is performed by penalizing the coefficients of the regression variables shrinking some to zero. The features that have non-zero coefficients are selected to be part of the model \cite{chandrashekar2014survey} \cite{powers2020evaluation}.

\subsubsection{Tree based method}
In \cite{powers2020evaluation} tree regularization framework was proposed to perform feature selection efficiently. The algorithm sequentially selects features which provide substantially new predictive information about the target variable. The relevant features selected are expected to contain informative but non-redundant features. The Tree method caters for both strong and weak classifiers. Another advantage of these methods is that they can deal with categorical and numerical, missing, and different scales between features and nonlinearities. 

\subsubsection{Self-organizing Maps}
Self-organizing Maps (SOMs) are used for projection of data in low dimensional spaces, usually 2-dimension (2-D), proposed by Kohonen in \cite{kohonen1990self}. This model consists of a set of \textit{C} discrete cells known as the "map". This map is made of a discrete topology which is defined by two dimensional graphs. A SOM architecture consists of two layers; the input and computational layers \cite{kamruzzaman2011new}. The input layer usually is made of the source nodes which represents the input features  \cite{kohonen1990self} \cite{kohonen1997self} \cite{lebbah2007besom}. According to \cite{kohonen1990self}, the determination of the set of weights (W) parameters is by minimizing the cost function iteratively as shown in equation \eqref{eq:7} .
\begin{equation}
	R(C,W)=\sum_{i=1}^{N}\sum_{j=1}^{|W|} K_{j,c(x)} ||X_{i}-W_{j}||^{2}
	\label{eq:7}
\end{equation}
Where $K_{j,c(x)(t)}$ is further expressed as;
\begin{equation}
	K_{j,c(x)(t)} =\exp (\frac{-\delta^{2}_{j,c}}{2\sigma^{_{2}}(t)}t=0,1,2,.....)
\end{equation}
for
\begin{equation}
	t=0,1,2,.....
\end{equation}
Where 
$ K_{j,c(x)(t)}$ is known as the neighbourhood function between each of the unit (j) on the map and the winning unit $C(x_{i})$ at the $t^{th}$ training step, $ \delta_{j,c(x)(t)}$ is known as the distance usually Euclidean, between unit (j) and the winning unit $C(x_{i})$ on the map and $\sigma$(t) represents the effective width of the topological neighbourhood at the training step $t^{th}$. The strength of SOMs are in their learning transparency. A SOM’s training is quite explainable and transparent. This can be done by assessing and studying the input mapping change at every epoch with the best matching unit (BMU) node \cite{kohonen1990self} \cite{lebbah2007besom}. The principal goal of a SOM is to transform an incoming signal pattern of arbitrary dimension into a one- or two-dimensional discrete map, and to perform this transformation adaptively in a topologically ordered fashion. A SOM can be set up by placing neurons at the nodes of a one- or two-dimensional lattice. One of the weaknesses of SOMs is that all input features are given the same level of relevance. During training, SOMs treat all features with equality and do not account for the contribution of the features after training, although further analysis of the SOM weights could reveal this. Another limitation is in its manual initialization of the number of nodes. A SOM requires a pre-defined size of SOM dimension as input parameter. This poses a problem when the underlying data complexity exceeds the SOM dimension or where there is no prior knowledge of the dynamic distribution of the inputs. To overcome the limitation of pre-defined size of SOM dimension, the Growing neural gas (GNG) was proposed in \cite{fritzke1995growing}. The idea behind the GNG is to grow the nodes by successively adding new nodes to an initially small network through evaluation of local statistical measures that were gathered during previous adaptation steps. The GNG grows during the training to find the appropriate size (number of nodes) for a given data \cite{fritzke1995growing}. According to \cite{fritzke1995growing}, the GNG has an advantage over the standard SOM in the sense that, in GNG the correct number of nodes is not expected to be decided a priori. For this reason, the GNG is suitable in cases of unknown distribution of sample observations. The key challenge of GNG is in the mechanism for deciding when to add a new node. The GNG adds nodes at constant number of intervals. In order to overcome this, a ceiling is usually provided to indicate when to stop adding (growing) nodes.

\subsection{Growing Self-Organising Maps}
The Growing Self-Organising Map (GSOM) proposed in \cite{alahakoon2000dynamic} has some characteristics similar to the GNG but uses a weight initialization method, which is simpler and reduces the possibility of twisted maps. The GSOM also carries out node generation interleaved with the self-organization, thus letting the new nodes smoothly join the existing network. The GSOM is an unsupervised neural network. Similar to the GNG, the GSOM grows nodes to represent the input data beginning with four nodes unlike GNG which starts with two nodes \cite{alahakoon2000dynamic}. During this node growing stage, the node weight values are self-organized similar to what is found in the SOM. The most interesting part of the GSOM is the growing phase. In this phase, a set of neuron weight vectors in the GSOM can be considered as a vector quantization of the input space such that each neuron is used to represent a region in which all points are closer to the node than the weight vector of any other neuron. In the growing phase, when input data are presented to the network, a winner is found and the difference between the input vector and the corresponding weight vector of the winner is accumulated as error value for that neuron. 
The error value calculated for each node can be considered as a quantization error and the total quantization error is represented in equation \eqref{eq:8}

\begin{equation}
	QE=\sum_{i=1}^{N}E_{i}
	\label{eq:8}
\end{equation}

where $N$ is the number of neurons in the network and $E_{i}$ is the error value for neuron $i$. The total quantization error $QE$ is used as a measure of determining when to generate a new neuron. If a neuron contributes substantially toward the total quantization error, then its Voronoi region $V_{i}$ in the input space is said to be under-represented by neuron $i$. Therefore, a new neuron is created to share the load of neuron $i$.

\section{Review of Related Works}
There have been some FS algorithms proposed by researchers. As inciteful and promising as some of these algorithms are, they are faced with various limitations. For instance, Paniri et al. have proposed a multi-label feature selection method using swarm intelligence ant colony optimisation (ACO) \cite{paniri2020mlaco}. Using the multi-level kNN (ML-kNN) classifier, the method showed a better performance over some FS algorithms. Also, the ACO has been applied as a feature selection method in financial crisis prediction \cite{uthayakumar2020financial}, in breast cancer detection \cite{saranya2020malignant} and in the assessment of humorous speeches by TED speakers \cite{adi2020assessment}.

Biswas et al. proposed a parameter independent fuzzy weighted k-Nearest neighbour classifier (PIFW-kNN) in \cite{biswas2018parameter}. This method embeds the FS ability into the kNN classifier. PIFW-kNN exposes the challenge of choosing a suitable value of k and a set of class dependent optimum weights for the features as a single-objective continuous non-convex optimization problem. The authors solved this problem by using a very competitive variant of Differential Evolution (DE), called Success-History based Adaptive DE (SHADE). They perform extensive experiments to demonstrate the improved accuracy of PIFW-kNN compared to the other state-of-the-art classifiers. However, this method cannot output the relevant features for each class.

Gao and his co-authors,  in their work "the dynamic change of selected feature with the class (DCSF)" method proposed a feature selection technique based on class-specific mutual information variation. The DCSF method considers the dynamic change of selected features with the class unlike the traditional feature selection methods. However, the DCSF outputs global relevant features for the entire dataset instead of a set of relevant features for each class \cite{gao2018class}.

A kNN method for lung cancer prognosis with the use of a genetic algorithm for feature selection has been proposed in \cite{maleki2021k}. In this method, a genetic algorithm is applied as feature selection to reduce the dataset dimensions. This method is employed in diagnosing or identifying the stage of patients’ disease. To improve the accuracy of the proposed algorithm, the authors opined that the best value for k is determined using an experimental procedure. The implementation of the proposed approach on a lung cancer database revealed 100$\%$ accuracy. This implies that the algorithm could be used to find a correlation between the clinical information and data mining techniques to support lung cancer staging diagnosis efficiently.

A new variant of GA named as the binary chaotic genetic algorithm (BCGA) showed an improvement over traditional GA in feature selection tasks using AMIGOS (A Dataset for Affect, Personality and Mood Research on Individuals and Groups) and two healthcare datasets having large feature space \cite{novelbinary2020}.

The performances of these methods have been found to be quite impressive as they are seen to have performed remarkably well in their areas of applications. However, it is observed that all the methods carry out global feature selection; that is, they select features that are common to all the classes. This will be a problem for any data where features are specific to individual classes, which is more common in real world data problems. There are, however, few non-metaheuristic algorithms which attempt at class-specific feature selection tasks \cite{sudha2012effective} \cite{rajesh2012analysis}. In \cite{ezenkwu2021class}, a class-specific metaheuristic technique for explainable relevant feature selection was proposed. Although the method attempted to identify the relevant features at class level, it can only identify the number of relevant features based on the suggested number of features provided by the user. 

In general practice, the feature selection methods assess their performances in terms of their prediction accuracies. As a result, we created some synthetic datasets and used them alongside popular real world datasets published on the UCI repository \cite{Dua:2019} to evaluate these FS methods. The reason for using synthetic datasets is to properly have the characteristics of class-specific feature relevance modelled and represented in the dataset, since this is not often known for real world datasets, and it is important to evaluate the FS methods against these characteristics so that their limitations can be fully understood. In this paper, we propose a class-level feature selection technique using the GSOM. We achieve this by applying feature weightings on the GSOM training information and carry out feature relevance analysis at both the class level and node level. The class level feature analysis reveals which features are relevant for which class while the node level feature analysis reveals the feature relevance for each node of the GSOM. Although attempts were made in \cite{starkey2017semi} and \cite{starkey2017automated} to achieve feature analysis using the feature weighted self organising maps FWSOM, it was however with many limitations. One of the limitations was the inability to choose the correct and appropriate map size for the dataset. This resulted in a trial-and-error approach of the choice of the map size. In our new method, this has been overcome as the GSOM automatically grows the node to represent or match the dataset appropriately. 

\section{Feature Weighted Growing Self organising Map}
Inspired by the automated node growing ability of the GSOM, and the earlier work on the FWSOM approach \cite{starkey2017automated},\cite{starkey2017semi}, we propose our method which uses the information from the GSOM to reveal feature relevance at class level, by controlling what features influence the separation of each class with a weighting function during training.  The key insight here is to note that any unsupervised learning mechanism makes decisions about its training data, and this can therefore be investigated in terms of its correctness, and modified if it is wrong if the algorithm permits such modification.
Exploiting this learning process enables it to select features which are relevant for each class. Our method, Feature weighted Growing Self-Organising Map (FWGSOM) consists of three stages viz; the training stage, the feature analysis stage and the classification stage. The FWGSOM method seeks to automatically identify what features are important in a dataset at class level in order to improve the classification accuracy. In this proposed approach, information from what the GSOM has learnt during training is used to identify what the GSOM sees as important for making decisions and to guide subsequent steps of the training.  For example, the SOM and GSOM approaches, once trained on a dataset, make decisions about the input data in an unsupervised manner.  A sample presented to the SOM or GSOM will result in a BMU being identified, and if all samples from the same class map to the same node, and samples from other classes to other nodes, then it can be said that the decision that the node makes in identifying these samples is correct for this class.  The node can then be examined against the other nodes to determine what this decision is, and is possible due to the transparent nature of these learning algorithms.   Conversely, if samples from a class do not map to a single node (or group of neighbouring nodes), then the decisions that are being made by the SOM or GSOM can be examined, with features considered to be irrelevant reduced in importance.  This process can be repeated until changes in the sample BMUs is seen, and the analysis can begin again.   This helps to generate individual weightings at a node level which will reduce the importance of inputs that are considered to be irrelevant for that node and class. It is expected that samples from a given class may spread over multiple nodes (i.e. will not be mapped to a single node) due to any irrelevant features in the dataset, and also due to noise in the dataset.  When there are irrelevant inputs in the dataset, the GSOM may see samples from the same class as different due to these irrelevant inputs, but the FWGSOM method will attempt to identify these features and reduce their importance so that the FWGSOM will instead see these samples as similar to each other. 

\subsection{FWGSOM Feature Relevance Analysis} 
In order to analyse the samples that map to any given node, the Samples Hit Matrix is computed once the GSOM has completed training.  The Samples Hit Matrix determines the number of samples from each class mapped to each node.  The Samples Hit Matrix can be visualised in the form of a pie chart, which shows the percentage of samples from each class in each node, usually distinguished with colours. An example of Samples Hit Matrix is shown in fig. \ref{fig:lungsom}, and shows analysis for a trained 3x3 SOM for Lung Data \cite{Dua:2019}. The Samples Hit Matrix in most cases is used to measure the classification accuracy of the SOM or GSOM. There are three groups of nodes for each individual class, namely, the class lead node (single node that is assumed to best represent the class at this point in the training), the class associates node (single or multiple nodes that are not lead nodes for the class but have samples from the class mapped, which may or may not be strictly neighbouring to the lead node for the class) and the class dissociate nodes (single or multiple nodes that do not have samples of that class mapped, and therefore it is assumed that these nodes are different from the class lead node). 
We carry out the analysis of these nodes by calculating the difference between the average of these nodes and the class lead node, and is possible due to the transparent nature of the learning of these unsupervised algorithms. 

\subsection{Similarity matrix} 
We calculate the similarity matrix for each class as the distance between the mean of the input samples $\bar{X}_{ck}^{n}$ that map to the nodes $N_{NbC_{k}}$ and the corresponding $N_{WnC_{j}}$ weight values in equation \eqref{eq:9}.

\begin{equation}
 Sim_{C_{k}}^{d}=||\bar{X}_{ck}^{n}-W_{N_{WnC_{j}}}^{d}||
 \label{eq:9}
\end{equation}

Where $n$ is the input dimension of a given sample $X$ belonging to class $C_{j}$  and $W$ is the weight value of the winner node $N_{WnC_{j}}$. 
The distance values in the similarity matrix are expected to be low for relevant features since it is assumed that these features will share similar values for the same class. For noiseless data, the values are expected to be exactly equal if a single relationship describes the class.  When the neighbouring nodes are connected or linked in the GSOM topological mapping, then this statement is self-evident but may not be true if the neighbouring nodes are spread out over the GSOM and are not adjacent to each other. Samples can be spread out over the GSOM due to noise in the data, irrelevant inputs or because there is more than one relationship that describes the class.  On the other hand, the distance values for irrelevant inputs are expected to be high since it is assumed that these features will have different values in the GSOM for the nodes to which the class samples are mapped and are the primary reason why the class samples are mapped to different nodes rather than a single node. This also assumes that the level of noise in the relevant features is less than the difference in values for irrelevant features within the class.

\subsection{Dissimilarity matrix} 
Similar to the similarity matrix, a dissimilarity matrix is assumed to describe the features that are different from the class under consideration and all other classes. This is computed as shown in equation \eqref{eq:10} as the distance between the mean of the class samples in the winner node $\bar{X}_{Nwnc_{j}}^{d}$ to the weight values of $N_{DistC_{j}}$ as;

\begin{equation}
	DS_{C_{j}}^{n}=||\bar{X}_{N_{wnC_{j}}}^{d}-W_{N_{DsC_{j}}}^{n}||
	\label{eq:10}
\end{equation}

This stems from the assumption that the GSOM did not map the samples from the class to these other distant nodes due to the inputs that the GSOM sees as different, and therefore the distance values in the dissimilarity matrix for these inputs are expected to be high.

We calculate the percentage change for each of the classes as expressed in equation \eqref{eq:11}.

\begin{equation}
	C_{k}\delta=(\frac{\sigma^{2}_{(DS_{C_{j}}^{n})}-\sigma^{2}_{(Sim_{C_{j}}^{dn})}}{\sigma^{2}_{(Sim_{C_{j}}^{dn})}})*100
	\label{eq:11}
\end{equation}

For each class, important inputs are identified as inputs with percentage change value greater than zero. 
After the identification of the relevant features, the weighting is then applied to each feature. In this work, we assign a weighting of 0 or 1 for features identified as irrelevant or relevant respectively, although it is also possible to instead reduce the weighting of a given feature instead, and thereby have an iterative process for finding the relevant features for a class.

An overview of the algorithm is given in Algorithm \ref{alg:FWGSOM}, with the main features of the algorithm as described in the preceding sections reflected in the pseudocode.

\begin{figure}
\caption{FWGSOM Pseudocode} \label{alg:FWGSOM}
\begin{algorithmic}[1]
\STATE User provides data, plus labels for classes
\STATE Train a GSOM using input data
\STATE Calculate class diagnosis accuracy
\WHILE{$iterations < MaxIter + class accuracy < TargetAccuracy$}
\STATE Analyse GSOM for Similarity matrix
\STATE Analyse GSOM for Dissimilarity matrix
\STATE Apply feature weighting at node level
\STATE Re-evaluate winning nodes for all data samples
\STATE Update class diagnosis accuracy
\ENDWHILE
\STATE Return feature weightings to user with attained classification accuracy for each class
\end{algorithmic}
\end{figure}

\section{Methodology}

\subsection{Descriptions}
In order to evaluate the performance of the FWGSOM and the FS methods, a number of experiments were designed and carried out. A problem with many stochastic machine learning algorithms is that different runs of the same algorithm on the same data may return
different results, although the variance may be negligible in some cases. This means that when performing experiments to assess the performance of an algorithm or to compare algorithms, it is usually a recommended practice to collect multiple results by repeating the experiment many times and the mean and standard deviation of the results are used to
summarize the performance of the model. Although, there is no standard number of times an experiment should be repeated, some researchers suggest minimum of 20 times. However, considering the size of the datasets for this research, which impacts on the execution time for each experiment, the experiments were repeated 15 times for the datasets on each algorithm. It is expected that each experiment may give different results. This explains the reason for the spread of outputs obtained in the results shown in the box plots.  The performance of the FWGSOM was compared to a number of FS methods, with feature selection accuracy as the main metric for datasets with known feature relevance and classification accuracies following the feature selection process for real world datasets using a number of classification engines for comparison purposes. The breakdown of the experiments and what they seek to achieve are discussed in the subsections below.

\subsubsection{Computational footprint performance} 
In this experiment, we computed the carbon footprint and running time for all of the methods discussed in the preceding sections.  For running time, the Blob$\_$D dataset was used since this is a more difficult dataset to analyse and was run with 5000 samples and an increasing number of features up to 1000 features.  For the carbon footprint calculations the final Blob$\_$D dataset of 5000 samples with 1000 features was used. This is an increasingly important consideration for data analysis methods, since there is an ethical imperative to consider the impact of computational footprint for the various methods being considered and should therefore form part of our considerations when comparing methods against each other, due to the impact of climate change. In doing this, we adopted the Green Algorithms Framework proposed in \cite{lannelongue2021green} to calculate the carbon footprint emissions of the FS and classification methods. The carbon footprint calculation is based on the processing time, type of computing cores, memory available, the efficiency and location of the computing facility. Mathematically, the energy consumption as presented in \cite{lannelongue2021green} is as expressed in equation \eqref{eq:12}. 
   
\begin{equation}
	E=t*(n_{c}*P_{c}*u_{c}+n_{m}*P_{m})*PUE*0.001
	\label{eq:12}
\end{equation}
where $t$ is the running time (hours), $nc$ the number of cores, and $n_{m}$ the size of memory available (gigabytes), $u_{c}$ is the core usage factor (between $0$ and $1$), $P_{c}$ is the power draw of a computing core, $P_{m}$ the power draw of memory (Watt) and $PUE$ is the efficiency coefficient of the data centre.
The carbon footprint $C$ (in $gCO_{2}e$) of producing a quantity of energy $E$ (in $kWh$) from sources with a $CI$ (in $gCO_{2}e$ $kWh^{-1}$) is expressed as \eqref{eq:13}

\begin{equation}
	C=E*CI
	\label{eq:13}
\end{equation}
With equations \eqref{eq:12} and \eqref{eq:13}, the carbon footprint can be expressed as \eqref{eq:14}
\begin{equation}
	E=t*(n_{c}*P_{c}*u_{c}+n_{m}*P_{m})*PUE*CI*0.001
	\label{eq:14}
\end{equation}

\subsubsection{Varying interclass distances}
This is an important part of the experiment and continues work examining the problem of varying interclass distance as extensively discussed in \cite{akpan2021review}. The problem describes a dataset whose classes are unequally separated, and are separated by different amounts. Table \ref{tab:d3distances} and Table \ref{tab:d4distances} show the dataset with equally separated (equal class distance) and varying interclass distance between class 1 and class 2 respectively. These datasets were tested on all methods to compare their performance.

\subsubsection{Global and Class-based features selection}
Another important part of this experiment is to examine how the various algorithms perform in terms of ability to carry out FS. As mentioned earlier in the literature review, most methods can only carry out global FS. This implies selection of features that are common to all classes. However, many real world datasets have features which are relevant for one class and noisy or irrelevant to another class. Therefore, it is important to identify those FS methods that are capable of undertaking FS at class level. 

\subsubsection{Classification performance}
As stated earlier, there are real world datasets whose information on feature relevance is unknown. In such cases, it is impossible to directly assess the performance of any FS method. The best and only means of assessing their performance is to evaluate them with respect to the classification accuracies using independent classifiers. If the correct features were selected, the classification accuracies of the classifiers are expected to improve accordingly. The classifiers used are SOM, Support vector machine (SVM) and Deep learning (DL), with the settings and hyperparameters used for all training as shown in Table \ref{tab:AIsettings}. The SVM and DL are supervised learning algorithms; they seek to optimise a cost function for the similarity between the model prediction and the ground truth using gradient descent methods. However, these methods can be both time and sample inefficient. Moreover, the architectures of these methods are incapable of supporting explainability since they are inherently black box approaches.

\begin{table*}
	\caption{Settings and hyperparameters used for classification algorithms}
	\label{tab:AIsettings}
	\begin{center}
		\begin{tabular}{| c | c |} 
			\hline
			Algorithm & Setting \\ 
			\hline
			GSOM & Spread factor=0.9\\
			& Number of iterations = 100\\
			& Initial learning rate=0.7\\
			\hline
			SOM & Som Size = [8 8]\\
			& Number of iterations = 100\\
			& Initial learning rate = 0.7 \\
			\hline
			SVM & Kernel function = Linear and cubic \\
			& Kernel scale = automatic\\
			& Box constraint level = 1\\
			& Multiclass method = One vs One\\
			& Standardized data = True\\
			& PCA = disabled \\
			\hline
			Deep learning & Number of hidden layers = 50-100\\
			& Number of neurons per layer = 100 \\
			& Initial learning rate = 0.1\\
			\hline						
		\end{tabular}
	\end{center}
\end{table*}

\subsection{Experimental dataset}
To achieve the purpose of the experiment, synthetic datasets with problems or characteristics of typical real world datasets were created. These include datasets with varying degrees of noise and most importantly with class specific feature relevance. The use of the synthetic datasets is necessary and important because they allow for a full assessment and evaluation of the selected algorithms on how well they perform, what data problem can be solved by them, and expose where they may encounter difficulties if any. 
Furthermore, synthetic datasets can exhibit the attributes of real-world datasets in terms of different data shapes, forms and presence of noise, unequal input variance, overlapping class definitions and class specific feature relevance etc. However, while the synthetic datasets can mimic many attributes of real-world datasets, they do not copy the original content of the real-world datasets exactly. As a result, some real world datasets from the UCI repository were also used to test the efficacy of the FWGSOM and other methods. The datasets are described in Table \ref{tab:table}.

\subsection{Data visualisation}
\begin{figure}
	\centering
	\includegraphics[width=1\linewidth]{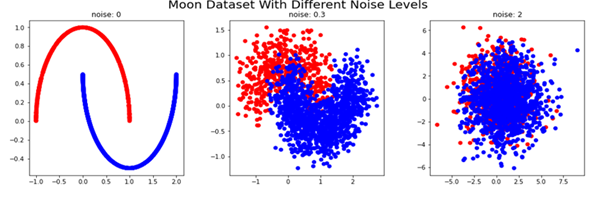}
	\caption{Visualisation for Moon$\_$D}
	\label{fig:moon}
\end{figure}

\begin{figure}
	\centering
	\includegraphics[width=1\linewidth]{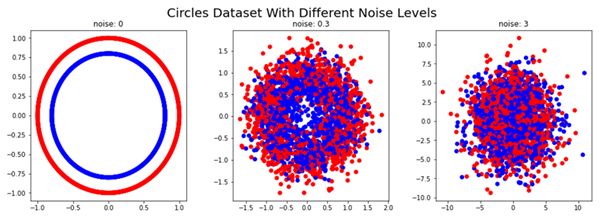}
	\caption{Visualisation for Circle$\_$D}
	\label{fig:circle}
\end{figure}

\begin{figure}
	\centering
	\includegraphics[width=1\linewidth]{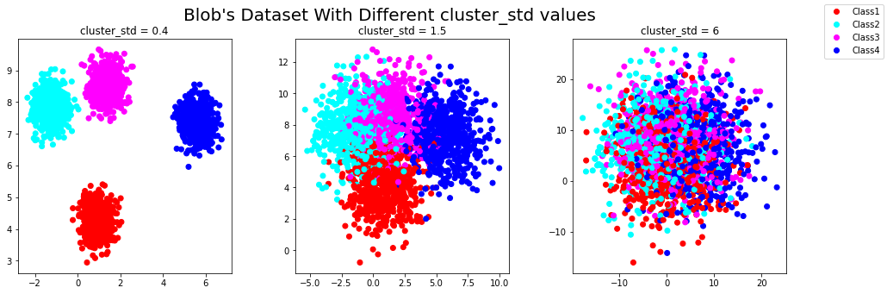}
	\caption{Visualisation for Blob$\_$D}
	\label{fig:blob}
\end{figure}

\begin{figure}
	\centering
	\includegraphics[width=1\linewidth]{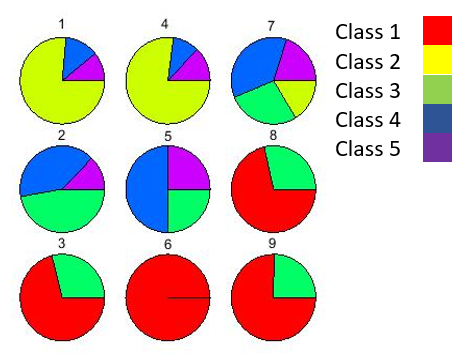}
	\caption{Lung Data Samples Hit Matrix for 3x3 SOM}
	\label{fig:lungsom}
\end{figure}

The data visualisation for some of the synthetic datasets with varying levels of noise (0\%, 30\% and 200\% noise) for the Moon$\_$D, Circle$\_$D and Blob$\_$D data shapes are shown in fig. \ref{fig:moon} , fig. \ref{fig:circle} and fig. \ref{fig:blob} respectively, and show varying degree of difficulty for the FS methods. Details of these datasets are described in \cite{scikit-learn} and \cite{sklearn_api}. The datasets can be obtained online as provided by \cite{scikit}.

%

\begin{table*}
	\caption{Datasets and Properties}
	\label{tab:table}
	\begin{tabularx}{\textwidth}{|l |X |}
		\hline
		Dataset Name & \textbf{Properties and Description}.\\
		\hline
		D$\_$1 & \textbf{Total samples: 2442    Features: 12    Classes:  5}\\&
		Relevant features are 1,2,3,4 for all classes. Other features are noisy features.
		\\  
		\hline
		D$\_$2 & \textbf{Total samples: 2442    Features: 28    Classes:  5}\\&
		Relevant features for Class 1 are 1, 2, 3 \\&
		Relevant features for Class 2 are 1, 2, 3 \\&
		Relevant features for Class 3 are 2, 4, 5 \\&
		Relevant features for Class 4 are 1, 3, 5, 6 \\&
		Relevant features for Class 5 are 1, 3, 4, 7 \\&
		Noise features are 8, 9, 10 to 28 
		\\
		\hline
		D$\_$3 & \textbf{Total samples: 2442    Features: 15    Classes:  5}\\&
		Relevant features for Class 1 are 1, 2, 3 \\&
		Relevant features for Class 2 are 4, 5, 6 \\&
		Relevant features for Class 3 are 2, 3, 4, 5 \\&
		Relevant features for Class 4 are 6, 7, 8 \\&
		Relevant features for Class 5 are 1, 4, 8 \\&
		Noise features are 9 to 15 
		\\
		\hline
		D$\_$4 & \textbf{Total samples: 1200    Features: 16    Classes:  6}\\&
		The dataset contains 16 features and 6 classes with equal distribution of 200 samples in each class. This is a clean dataset with 4 irrelevant inputs and unequal class separations (varying inter class distances). It was created by reducing the class distance between class 1 and class 2 by 200$\%$ while other distances distances were approximately equal, as shown in Table \ref{tab:d3distances} and Table \ref{tab:d4distances}.
		\\
		\hline
		Moon$\_$D & \textbf{Total samples: 2500    Features: 20    Classes:  2}\\&
		Relevant features are 1 and 2 for all classes. Other features are noisy features.
		\\
		\hline
		Circle$\_$D & \textbf{Total samples: 1750    Features: 20    Classes:  2}\\&
		This is an imbalanced data set with relevant features of 1 and 2 for all classes. Other features are noisy features.
		\\
		\hline
		Blob$\_$D & \textbf{Total samples: 1385    Features: 20    Classes:  4}\\&
		Relevant features are 1,2,3,4,5,6. Others are noisy features of different degrees
		\\
		\hline
		Waveform Data & \textbf{Total samples: 5000    Features: 40    Classes:  3}\\&
		This is a real-world dataset from UCI repository. As described on UCI platform, the first 21 inputs of the waveform data describe the classes, the latter 19 are completely irrelevant and noise features with mean 0 and variance 1. No information is provided on class specific relevance for the 21 relevance features.
		\\
		\hline
		Glass Identification Data & \textbf{Total samples: 214    Features: 10    Classes:  6}\\&
		This dataset is from USA Forensic Science Service obtained from the UCI repository \cite{Dua:2019}. There is no information on feature relevance.
		\\
		\hline
		Lung Data & \textbf{Total samples: 203    Features: 3312    Classes:  5}\\&
		The LUNG dataset was obtained from the UCI repository \cite{Dua:2019}. The information on feature relevance is not provided.
		\\
		\hline
		
	\end{tabularx}
\end{table*}

\begin{table}
	\caption{Average distances between classes for D$\_$3: Equally separated classes}
	\label{tab:d3distances}
	\centering
	\begin{tabular}{|c |c c c c c c|}
		\hline		
		$C_6$ & 80.2 & 64.4 & 47.9 & 32.4 & 16.6 & 0 \\
		
		$C_5$ & 63.6 & 47.9 & 31.4 & 15.9 & 0 & 16.6\\
		
		$C_4$ & 47.8 & 32.1 & 15.6 & 0 & 15.9 & 32.4 \\
		
		$C_3$ & 32.3 & 16.5 & 0 & 15.9 & 31.4 & 47.9 \\
		
		$C_2$ & 15.8 & 0 & 16.5 & 32.1 & 47.9 & 64.4 \\
		
		$C_1$ & 0 & 15.8 & 32.3 & 47.8 & 63.6 & 80.2 \\
		\hline
		& $C_1$ & $C_2$ & $C_3$ & $C_4$ & $C_5$ & $C_6$ \\
		\hline
	\end{tabular}
\end{table}

\begin{table}
	\caption{Average distances between classes for D$\_$4: Unequally separated classes}
	\label{tab:d4distances}
	\centering
	\begin{tabular}{|c| c c c c c c|}
		\hline	
		$C_6$ & 72.4 & 94.4 & 51.8 & 34.2 & 18.7 & 0 \\
		
		$C_5$ & 55.8 & 52.8 & 35.2 & 17.6 & 0 &  16.7 \\
		
		$C_4$ & 38.2 & 35.2 & 17.6 & 0 & 17.6 & 34.2 \\
		
		$C_3$ & 20.6 & 17.6 & 0 & 17.6 & 35.2 & 51.8 \\
		
		$C_2$ & 3.2 & 0 & 17.6 & 35.2 & 52.8 & 69.4 \\
		
		$C_1$ & 0 & \textbf{3.2} & 20.6 & 38.2 & 55.8 & 72.4 \\
		\hline
		& $C_1$ & $C_2$ & $C_3$ & $C_4$ & $C_5$ & $C_6$ \\
		\hline
	\end{tabular}
\end{table}

\section{Results and discussions}
In this section, we present the results of the performance of FWGSOM and other FS algorithms in terms of feature selection accuracy for the datasets with known feature relevance. For the real-world datasets with unknown feature relevance, the performance of the FS methods was evaluated by applying the selected features on some classifiers to assess any improvement in classification accuracies of the classifiers with the selected features as inputs. The classification accuracy is expected to improve if the relevant features in the dataset are selected correctly, since the noisy and/or irrelevant features will have been removed. These results are presented and discussed in this section also.

\subsection{Global feature datasets performance results}
\begin{figure}
	\centering
	\includegraphics[width=1\linewidth]{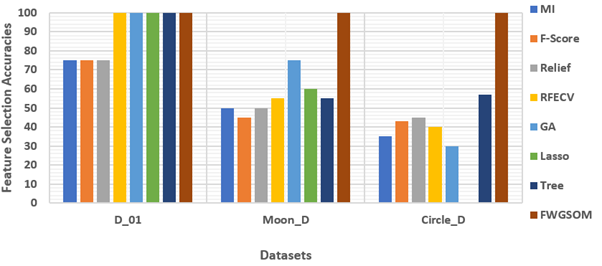}
	\caption{Feature selection performance for D$\_$1, Moon$\_$D and Circle$\_$D Datasets}
	\label{fig:fig6}
\end{figure}

This section discusses results for datasets where the classes have their relationship described by the same features.
Fig. \ref{fig:fig6} shows the FS performance for D$\_$1, Moon$\_$D and Circle$\_$D Datasets. While the RFECV, GA, Lasso and Tress all have 100$\%$ accuracies for D$\_$1, they all perform poorly on Moon$\_$D and Circle$\_$D datasets. The FWGSOM method shows superior performance on all the datasets with 100$\%$ FS accuracy.

\begin{figure}
	\centering
	\includegraphics[width=1\linewidth]{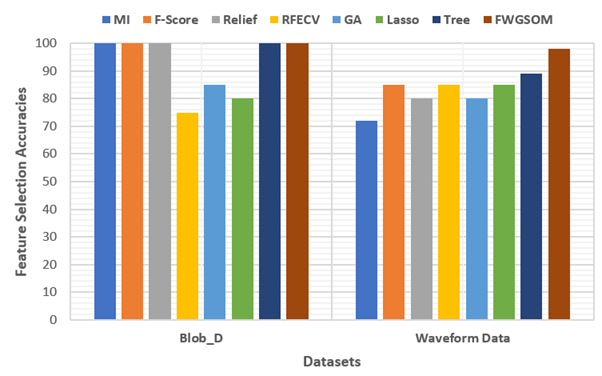}
	\caption{Feature selection performance for Blob$\_$D and Waveform Datasets}
	\label{fig:fig7}
\end{figure}

In fig. \ref{fig:fig7}, MI, F-Score, Relief, Tree and FWGSOM all have 100$\%$ FS accuracies on Blob$\_$D datasets. However, none of the methods has 100$\%$ accuracy on the Waveform Dataset, though the FWGSOM has the best performance of 98$\%$. Further investigation on the Waveform dataset reveals that the FWGSOM results indicate that one of the features is only relevant for some classes. Though this revelation appears valid based on the results obtained, it cannot be further confirmed, since it is a real world dataset and we can only rely and be guided by the information provided about the dataset on the UCI repository.

\subsection{Class-based feature datasets performance results}
This section discusses results for datasets where the classes use features that can be different from each other (i.e. the classes are not described by the same features).  The results of the experiments on FS are shown in Table \ref{tab:tabled2} and Table \ref{tab:tabled3}, for datasets D$\_$2 and D$\_$3 respectively which show the results of the FS approaches for the different features for different classes problem with the bold text indicating that the FS method has 100$\%$ selection accuracy. In Table \ref{tab:tabled2}, the FWGSOM is seen to have demonstrated excellent performance as it is able to select all the relevant features for each class. Other algorithms selected the correct features for the classes with addition of features not relevant to the classes. This proves their inability to carry out class specific FS. In a similar vein, Table \ref{tab:tabled3} show the FS results for dataset D$\_$3 which focussed on the same problem but with fewer features overall but with features being relevant for fewer classes. As can be observed in the results, similar to the results for D$\_$2, the FWGSOM once again outperforms other algorithms by selecting relevant features specific for each class in D$\_$3 with 100$\%$ accuracy. 

\subsection{Classification performance Results}
As stated earlier, for the datasets with no information on feature relevance, the performance of the different feature selection techniques were evaluated with respect to the classification accuracies using independent classifiers. The classifiers used are SOM, Support vector machine (SVM) and Deep learning (DL).

\begin{figure}
	\centering
	\includegraphics[width=1\linewidth]{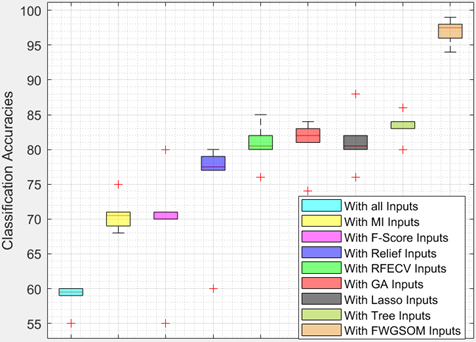}
	\caption{Classification performance of SVM using the FS methods on Glass dataset}
	\label{fig:svmglass}
\end{figure}

\begin{figure}
	\centering
	\includegraphics[width=1\linewidth]{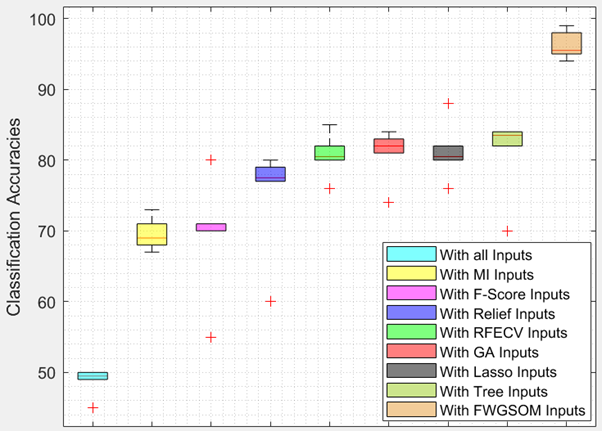}
	\caption{Classification Performance of SOM using the FS methods on Glass Dataset}
	\label{fig:somglass}
\end{figure}

\begin{figure}
	\centering
	\includegraphics[width=1\linewidth]{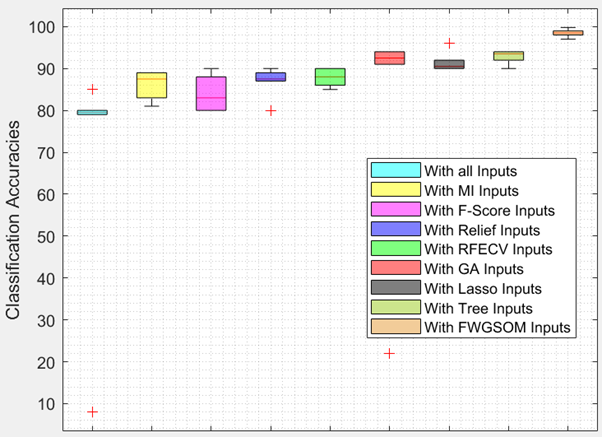}
	\caption{Classification Performance of DL using the FS methods on Glass Dataset}
	\label{fig:dlglass}
\end{figure}

\begin{figure}
	\centering
	\includegraphics[width=1\linewidth]{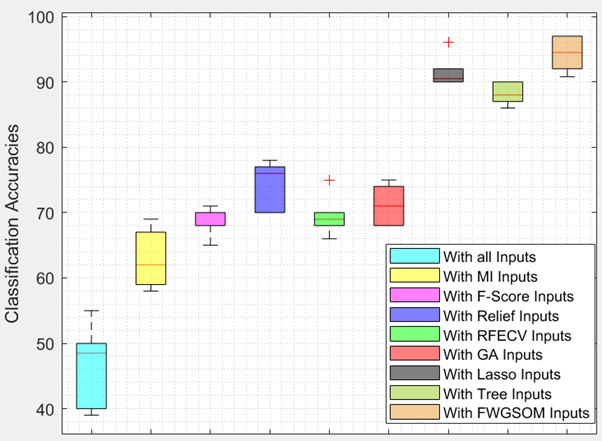}
	\caption{Classification Performance of SVM using the FS methods on Lung Dataset}
	\label{fig:svmlung}
\end{figure}

\begin{figure}
	\centering
	\includegraphics[width=1\linewidth]{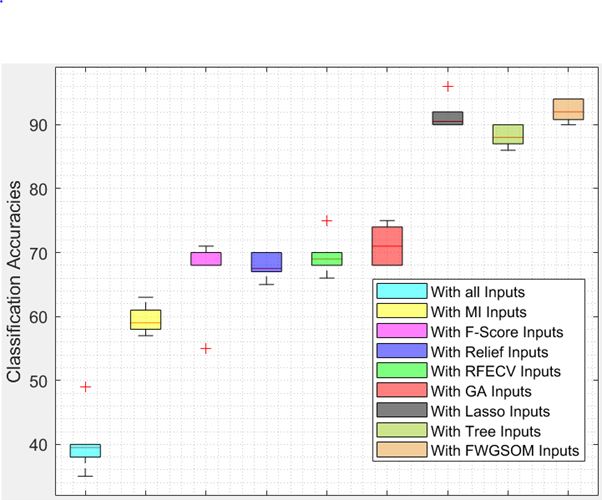}
	\caption{Classification Performance of SOM using the FS methods on Lung Dataset}
	\label{fig:somlung}
\end{figure}

\begin{figure}
	\centering
	\includegraphics[width=1\linewidth]{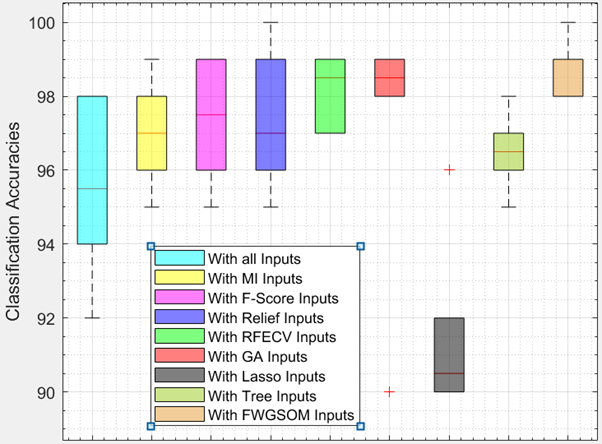}
	\caption{Classification Performance of DL using the FS methods on Lung Dataset}
	\label{fig:dllung}
\end{figure}

Figs \ref{fig:svmglass}, \ref{fig:somglass}, \ref{fig:dlglass}, \ref{fig:svmlung}, \ref{fig:somlung} and \ref{fig:dllung} show the results of the experiment on classification on the datasets with unknown feature relevance. From the results of the experiment, it can be inferred that all the classifiers have better classification performances when the inputs selected by any of the FS methods are used, but that the inputs selected by the FWGSOM approach consistently show the best improvement in performance for any of the classification methods. It is worth noting that there was no significant improvement with the DL classification performance when fed with the inputs from the FS methods on some datasets notably the Lung dataset. This can be explained by the ability of DL to automatically select the features itself and thereby to suppress noise and identify internally the relevant features in a dataset; thus enabling high classification accuracy. However an improvement in performance even with DL can still be seen when the FWGSOM selected features only are used for training. On the other hand, the SVM and SOM demonstrated significant improvement in classification when fed with inputs from the FS methods for all the datasets. These results are significant since they demonstrate that competitive performance from any classifier can be achieved if the correct features are identified for the types of problems studied in this paper.  Although the DL can obtain high accuracy of classification with datasets with various noisy features, the black box nature of DL means that the important features are not returned to the user which is an increasingly important topic particularly with a focus on XAI methods. Not surprisingly, the computational footprint for DL was seen to be the highest compared to the other classification methods as shown in Table \ref{tab:carbonfootprintAI}.


\begin{figure}
	\centering
	\includegraphics[width=1\linewidth]{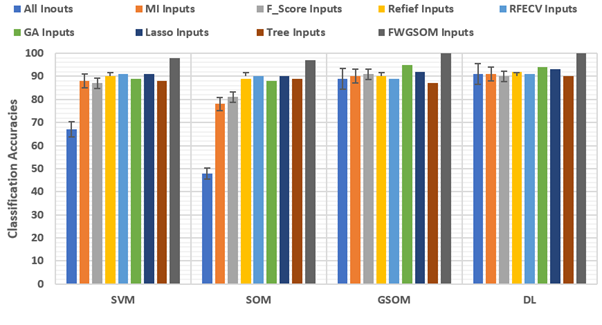}
	\caption{Classification accuracies for D$\_$04: Varying interclass distances Case I}
	\label{fig:alld04}
\end{figure}

%
%

\subsection{Performance of datasets with varying interclass distances}
The results of our experiments in \cite{akpan2021review} exposed the inability of classification algorithms to perform classification on datasets with varying interclass distances with high accuracies and that even DL methods are affected by this issue. The results in fig. \ref{fig:alld04} show that FWGSOM method can deal with datasets with varying interclass distances and correctly identify the relevant features for each class. It can be observed that when the inputs selected by FWGSOM are fed to the classification methods, there are tremendous improvements in the classification accuracies with GSOM and DL both achieving 100$\%$ accuracy.

\subsection{Computational footprint performance results} 
\begin{figure}
	\centering
	\includegraphics[width=1\linewidth]{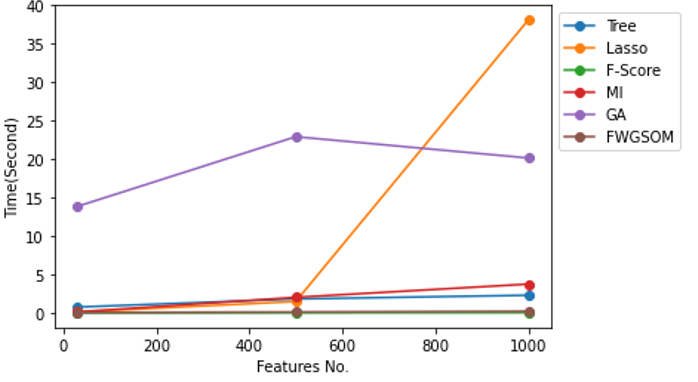}
	\caption{Running time for the feature selection methods}
	\label{fig:runningtime}
\end{figure}

This research also investigates the computational footprints of the proposed method and compares it with some existing methods for selected datasets. The metrics under consideration here are the running time and carbon footprints. The result in Fig. \ref{fig:runningtime} shows that the FWGSOM and F-Score methods have the shortest running time. This figure does not include the running time for Relief and RFECV methods as they are significantly higher and mean the differences in the lower values would not be seen.  In Table \ref{tab:carbonfootprintFS}, the FWGSOM records along with F$\_$Score and Relief the lowest carbon footprint of 0.0g$C0_{2}$e for all of the datasets.

\begin{table*}
	\caption{Feature selection results for D$\_$2}
	\label{tab:tabled2}
	\begin{center}
		\begin{tabular}{| c | c c c c | c c c c | c c c c | c c c c |} 
			\hline
			& \multicolumn{16}{c |}{FS method}\\
			\hline
			&	 \multicolumn{4}{c |}{MI}& \multicolumn{4}{c |}{F-Score} & \multicolumn{4}{c |}{Relief} &  \multicolumn{4}{c |}{RFECV}\\
			\hline
			
			&	SF &CSF &NF &AF&SF &CSF &NF &AF&SF &CSF &NF &AF& SF &CSF &NF &AF\\ 
			
			$C_{1}$ & 7 &3&4 &0&6	&2&	4&	0&7&	3 & 2&1&8 &3 &4 &1\\
			
			$C_{2}$ & 7 &2 &4&1&	6&3	&3&	0&	7 &3 &2&1 &8 &2 &4&2\\
			
			$C_{3}$ & 7 &2& 4&1&6	&3&	3&0&7 &3 &2&1 &8 &3&4 &1\\
			
			$C_{4}$ & 7 &4 &3&0	&6&	3&	2&1&	7 &4 &2 &1&8 &3&3 &2\\
			
			$C_{5}$ & 7 &4& 3&0&6&3&2	&1&	7 &4&3 &0 &8 &3&4 &1\\
			\hline
			\hline
			&	 \multicolumn{4}{c |}{GA}& \multicolumn{4}{c |}{Lasso} & \multicolumn{4}{c |}{Tree} &  \multicolumn{4}{c |}{FWGSOM}\\
			\hline
			&	SF &CSF &NF &AF&SF &CSF &NF &AF&SF &CSF &NF &AF& SF &CSF &NF &AF\\ 
			
			$C_{1}$ & 7 &2&4 &1&8	&3&	4&	1&8&	3 & 4&1&\textbf{3} &\textbf{3} &\textbf{0} &\textbf{0}\\
			
			$C_{2}$ & 7 &3 &4&0&	8&3	&3&	2&	8 &3 &4&1 &\textbf{3} &\textbf{3} &\textbf{0} &\textbf{0}\\
			
			$C_{3}$ & 7 &2& 4&1&8	&3&	3&2&8 &3 &4&1 &\textbf{3} &\textbf{3} &\textbf{0} &\textbf{0}\\
			
			$C_{4}$ & 7 &3 &3&1	&8&	4&	3&1&	8 &3 &4 &1&\textbf{4} &\textbf{4} &\textbf{0} &\textbf{0}\\
			
			$C_{5}$ & 7 &3& 3&1&8&4&3	&1&	8 &4&4 &0 &\textbf{4} &\textbf{4} &\textbf{0} &\textbf{0}\\
			\hline

		\end{tabular}
	\end{center}
	\footnotesize{SF=Selected Features: CSF=Number of correctly selected features: NF=Number of Noisy Features selected: AF=Number of additional features (Relevant for other classes) selected}
\end{table*}

%
%
%
%
%
%
%

\begin{table*}
	\caption{Feature selection results for D$\_$3}
	\label{tab:tabled3}
	\begin{center}
		\begin{tabular}{| c | c c c c | c c c c | c c c c | c c c c |} 
			\hline
			& \multicolumn{16}{c |}{FS method}\\
			\hline
			&	 \multicolumn{4}{c |}{MI}& \multicolumn{4}{c |}{F-Score} & \multicolumn{4}{c |}{Relief} &  \multicolumn{4}{c |}{RFECV}\\
			\hline
			
			&	SF &CSF &NF &AF&SF &CSF &NF &AF&SF &CSF &NF &AF& SF &CSF &NF &AF\\ 
			
			$C_{1}$ & 7 &3&3 &1&6	&2&	4&	0&7&	3 & 3&0&8 &3 &4 &1\\
			
			$C_{2}$ & 7 &2 &3&2&	6&3	&2&	1&	7 &3 &2&1 &8 &2 &3&3\\
			
			$C_{3}$ & 7 &2& 3&2&6	&3&	2&1&7 &3 &2&1 &8 &3&4 &1\\
			
			$C_{4}$ & 7 &4 &3&0	&6&	3&	2&1&	7 &4 &3 &0&8 &3&3 &2\\
			
			$C_{5}$ & 7 &4& 3&0&6&3&2	&1&	7 &4&3 &0 &8 &3&4 &1\\
			\hline
			\hline
			&	 \multicolumn{4}{c |}{GA}& \multicolumn{4}{c |}{Lasso} & \multicolumn{4}{c |}{Tree} &  \multicolumn{4}{c |}{FWGSOM}\\
			\hline
			&	SF &CSF &NF &AF&SF &CSF &NF &AF&SF &CSF &NF &AF& SF &CSF &NF &AF\\ 
			
			$C_{1}$ & 7 &2&3 &2&8	&3&	3&	2&8&	3 & 4&1&\textbf{3} &\textbf{3} &\textbf{0} &\textbf{0}\\
			
			$C_{2}$ & 7 &3 &4&0&	8&3	&3&	2&	8 &3 &2&3 &\textbf{3} &\textbf{3} &\textbf{0} &\textbf{0}\\
			
			$C_{3}$ & 7 &2& 4&1&8	&3&	3&2&8 &3 &4&1 &\textbf{3} &\textbf{3} &\textbf{0} &\textbf{0}\\
			
			$C_{4}$ & 7 &3 &2&2	&8&	4&	2&2&	8 &3 &3 &2&\textbf{4} &\textbf{4} &\textbf{0} &\textbf{0}\\
			
			$C_{5}$ & 7 &3& 2&2&8&4&3	&1&	8 &4&4 &0 &\textbf{4} &\textbf{4} &\textbf{0} &\textbf{0}\\
			\hline
		
		\end{tabular}
	\end{center}
	\footnotesize{SF=Selected Features: CSF=Number of correctly selected features: NF=Number of Noisy Features selected: AF=Number of additional features (Relevant for other classes) selected}
\end{table*}

\begin{table*}
	\caption{The Carbon Footprint (CO2 Equivalent) for the feature selection methods}
	\label{tab:carbonfootprintFS}
	\begin{center}
		\begin{tabular}{c c c c c c c c c} 
			\hline
			Datasets &	MI &	F$\_$Score & Relief &	RFECV &	GA &	Lasso &	Trees &	FWGSOM \\ [0.5ex] 
			\hline\hline
			D$\_01$&	 \textbf{0.0g$C0_{2}$e} &\textbf{0.0g$C0_{2}$e}&	\textbf{0.0g$C0_{2}$e}&	0.05g$C0_{2}$e&	0.11g$C0_{2}$e&	\textbf{0.0g$C0_{2}$e}&	0.01g$C0_{2}$e&	\textbf{0.0g$C0_{2}$e}\\ 
			\hline
			Blob$\_$D & 0.02g$C0_{2}$e&	\textbf{0.0g$C0_{2}$e}&8g$C0_{2}$e&	6g$C0_{2}$e&	0.10g$C0_{2}$e&	0.02g$C0_{2}$e&	0.02g$C0_{2}$e&	\textbf{0.0g$C0_{2}$e} \\
			\hline
			Blob$\_$D & 0.04g$C0_{2}$e&	\textbf{0.0g$C0_{2}$e}&7.6g$C0_{2}$e&	8.13g$C0_{2}$e&	0.12g$C0_{2}$e&	0.56g$C0_{2}$e&	0.02g$C0_{2}$e&	\textbf{0.0g$C0_{2}$e}\\
			\hline
			
		\end{tabular}
	\end{center}
\end{table*}

%

\begin{table*}
	\caption{The Carbon Footprint (CO2 Equivalent) for the classification methods}
	\label{tab:carbonfootprintAI}
	\begin{center}
		\begin{tabular}{c c c c c} 
			\hline
			Datasets &	SVM &	SOM & GSOM &	DL \\ [0.5ex] 
			\hline\hline
			Waveform Data&	0.9g$C0_{2}$e&	0.03g$C0_{2}$e&0.09g$C0_{2}$e&	20.14g$C0_{2}$e \\ 
			\hline
			Lung Data & 0.77g$C0_{2}$e&	0.14g$C0_{2}$e&0.17g$C0_{2}$e&	21.38g$C0_{2}$e \\
			\hline
			Glass Identification & 0.1g$C0_{2}$e &	\textbf{0.0g$C0_{2}$e}&\textbf{0.0g$C0_{2}$e}&	26.11g$C0_{2}$e\\
			\hline
			
		\end{tabular}
	\end{center}
\end{table*}

\section{Conclusion}
In this paper, a class-based FS method has been presented. This method called FWGSOM is inspired by the GSOM’s ability to automatically grow nodes to match the dataset characteristics. The FWGSOM’s ability to carry out feature analysis at class level enhances its ability to identify relevant features for each class. This overcomes the major limitation of global feature selection algorithms which select features as common to all classes. It has been observed that the global FS methods have shown substantial improvements in the predictive powers of classifiers for the datasets used. However, it has also been observed that they lack the capacity of addressing the problem of explainability of prediction outcomes. With the ability of FWGSOM to identify relevant features for each class in a dataset, we can conclude that our proposed method represents an improvement on global FS methods and that it outperforms all global FS techniques used in this paper. This has been demonstrated by its 100$\%$ feature selection accuracy for datasets with known feature relevance and significant improvement of classification accuracies for real world datasets with no information on feature relevance. Another significant output of this study is the finding that the FWGSOM is effective with datasets having varying interclass distances. Finally, but of increasing importance in today's world, is the finding that the FWGSOM was seen to be very effective and efficient in its computational carbon footprint. It is interesting to add the FWGSOM has the potential to automatically undertake classification following the feature selection phase. Further work in this area will also compare the computational footprint and XAI against other methods.

%

\bibliographystyle{IEEEtran}
\bibliography{ref}
\end{document}